\documentclass[runningheads]{llncs}
\usepackage[T1]{fontenc}
\usepackage{graphicx}
\graphicspath{{figures/}}

\usepackage{multirow}
\usepackage{booktabs}
\usepackage{amsmath,amssymb,amsfonts}
\usepackage{xcolor}
\usepackage{cite}
\usepackage{xspace}
\usepackage{bm}
\usepackage{url}
\usepackage[caption=false]{subfig}
\usepackage[rflt]{floatflt}

\usepackage{tcolorbox}
\newtcolorbox{answerbox}{
    colback=white,
    colframe=black!30,
    boxrule=0.8pt,
    arc=1pt,
    left=8pt,
    right=8pt,
    top=6pt,
    bottom=6pt
    }

\makeatletter
\DeclareRobustCommand\onedot{\futurelet\@let@token\@onedot}
\def\@onedot{\ifx\@let@token.\else.\null\fi\xspace}
 
\def\eg{{\it e.g}\onedot} 
\def\ie{{\it i.e}\onedot} 
 
\def\etc{{\it etc}\onedot}

\makeatother

\begin{document}
\title{Beyond Standard Benchmarks: A Systematic Audit of Vision-Language Model's Robustness to Natural Semantic Variation Across Diverse Tasks}
\titlerunning{Vision-Language Model Evaluation}
%
\author{Jia Chengyu\inst{1} \and
AprilPyone MaungMaung\inst{2} \and
Huy H. Nguyen\inst{2} \and \\
Jinyin Chen\inst{1} \and
Isao Echizen\inst{2}}
\authorrunning{J. Chengyu et al.}
\institute{Zhejiang University of Technology, Hangzhou, Zhejiang, China \and
National Institute of Informatics, Tokyo, Japan}

\maketitle              
\begin{abstract}
Recent advances in vision-language models (VLMs) trained on web-scale image-text pairs have enabled impressive zero-shot transfer across a diverse range of visual tasks. However, comprehensive and independent evaluation beyond standard benchmarks is essential to understand their robustness, limitations, and real-world applicability. This paper presents a systematic evaluation framework for VLMs under natural adversarial scenarios for diverse downstream tasks, which has been overlooked in previous evaluation works.
We evaluate a wide range of VLMs (CLIP, robust CLIP, BLIP2, and SigLIP2) on curated adversarial datasets (typographic attacks, ImageNet-A, and natural language-induced adversarial examples).
We measure the natural adversarial performance of selected VLMs for zero-shot image classification, semantic segmentation, and visual question answering.
Our analysis reveals that robust CLIP models can amplify natural adversarial vulnerabilities, and CLIP models significantly reduce performance
for natural language-induced adversarial examples.
Additionally, we provide interpretable analyses to identify failure modes.
We hope our findings inspire future research in robust and fair multimodal pattern recognition.

\end{abstract}

\section{Introduction}
Vision-language models (VLMs) represent a profound shift from a task-specific paradigm to an open-world, task-agnostic perception.
Concretely, the contrastive language-image pre-training (CLIP) model leverages contrastive training to align images and text within a unified embedding space, exhibiting remarkable versatility in zero-shot generalization across diverse tasks without task-specific adaptation~\cite{radford2021learning,jia2021scaling}. Notably, it demonstrates robust classification performance under natural distribution shifts~\cite{recht2019imagenet,wang2019learning,hendrycks2021many,barbu2019objectnet,hendrycks2021natural}. Beyond standalone applications, CLIP's image encoder serves as a foundational component in modern large vision-language models (LVLMs), including OpenFlamingo~\cite{awadalla2023openflamingo} and LLaVA~\cite{liu2023visual}, where it is integrated with large language models (\eg, MPT~\cite{mosaicml2023introducing} and Vicuna~\cite{chiang2023vicuna}). These LVLMs exhibit exceptional zero-shot capabilities in tasks such as image captioning~\cite{mokady2021clipcap} and visual question answering (VQA)~\cite{shenmuch}.

Given the growing adoption of LVLMs in real-world applications, ensuring their safety and alignment is critical.
Recent studies reveal their vulnerability to adversarial attacks, particularly in the visual modality~\cite{zhao2023evaluating,zhou2022enhancing}. For instance, even commercial systems like Google's BARD can be compromised by carefully crafted perturbations~\cite{dong2023robust}.
Researchers have proposed unsupervised adversarial fine-tuning methods (\eg, robust CLIP) to mitigate such threats~\cite{schlarmann2024robust}.
However, VLMs face new threats. Robustness is no longer limited to invariance against pixel-level corruptions, but also requires alignment with the wide range of semantic concepts expressed through language.
For example, typographic attacks, where overlaid text (\eg, a ``dog'' label on a cat image) misleads predictions~\cite{goh2021multimodal,materzynska2022disentangling,ilharco2022patching}.

\begin{floatingfigure}{.46\linewidth}
  \includegraphics[width=.44\linewidth]{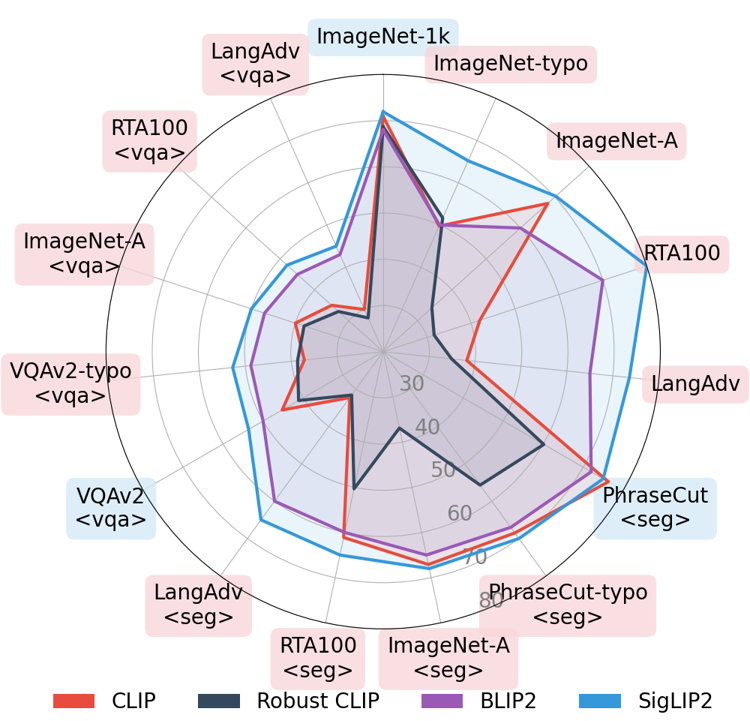}
  \caption{Performance comparison of vision-language models under natural adversarial scenarios. Large areas indicate better overall performance.}\label{fig:radar}
\end{floatingfigure}
Accordingly, understanding the capabilities, failure modes, and biases of VLMs such as CLIP is crucial for their responsible and effective deployment.
Therefore, a recent work evaluates CLIP for multiple robustness dimensions such as visual factor variations, uncertainty calibration, out-of-distribution detection, and modality interaction~\cite{tu2025toward}.
However, the previous evaluation effort falls short for two reasons. (a) There is limited natural adversarial robustness evaluation (only ImageNet-A~\cite{hendrycks2021natural} was used although there are other natural adversarial scenarios such as typographic attacks). (b) The focus is on image classification only although VLMs are used for diverse downstream tasks.

To fill this gap, we propose an evaluation framework for VLMs under natural adversarial scenarios for three downstream tasks: zero-shot image  classification, semantic segmentation, and visual question answering.
Our evaluation results reveal that robust CLIP models can amplify natural adversarial vulnerabilities (a previously unknown and serious risk) as shown in Fig.~\ref{fig:radar}, although ImageNet-1k classification is comparable.
CLIP models significantly reduce performance for natural language–induced adversarial semantics (overlooked in prior evaluations).
SigLIP2 models achieve overall best performance in natural adversarial settings.
Our contributions are as follows.
We construct adversarial datasets with (1) typographic attacks, where superimposed text misleads classification~\cite{goh2021multimodal}, (2) natural adversarial examples, comprising unmodified yet challenging real-world images (ImageNet-A~\cite{hendrycks2021natural}), and (3) natural language-induced adversarial images~\cite{zhu2024natural}, which exploit multimodal alignment through semantically aligned but misleading visuals.
We evaluate a wide range of VLMs, totaling 22 models (OpenAI CLIP~\cite{radford2021learning}, timm CLIP with a ResNet backbone~\cite{rw2019timm}, robust CLIP~\cite{schlarmann2024robust}, SigLIP2~\cite{tschannen2025siglip}, PaLiGemma2, and BLIP2~\cite{li2023blip}) under natural adversarial scenarios for three downstream tasks.
Beyond empirical benchmarking, we provide interpretable analyses to uncover failure modes: (i) CLIP's vision encoder overly relies on local textures (\eg, text artifacts) over global semantics, and (ii) its alignment mechanism is susceptible to spurious correlations between visual and textual features.

\section{Related Work}
\noindent{\bf Vision-Language Pretraining.}
Recent vision-language models (\eg, CLIP~\cite{radford2021learning}, ALIGN~\cite{jia2021scaling}, SigLIP2~\cite{tschannen2025siglip}, and BLIP2~\cite{li2023blip}) leverage large-scale image--text pretraining to enable strong zero-shot and transfer performance across a wide range of vision and multimodal tasks. CLIP-style models are typically trained using contrastive objectives on image--text pairs, while more recent architectures such as BLIP2 combine frozen vision encoders with lightweight language models to support both discriminative and generative tasks.
Beyond standard CLIP, robust CLIP~\cite{schlarmannrobust} has explored robustness-enhanced variants obtained by fine-tuning CLIP models on perturbation-based or adversarially augmented data. These models aim to improve resilience to input perturbations while retaining CLIP’s transfer capabilities.
However, the robustness of both standard and robustness-enhanced vision-language models is not yet rigorously evaluated under natural adversarial settings, especially for different downstream tasks.

\vspace{2mm}\noindent{\bf Robustness Evaluation.}
Prior work has developed several frameworks to assess the robustness of CLIP and related vision–language models. Early work primarily examined distribution shifts and adversarial perturbations in image classification, often using corrupted~\cite{hendrycks2018benchmarking} or out-of-distribution variants of ImageNet~\cite{hendrycks2021natural}. 
Tu et al.~\cite{tu2025toward} advocate for a holistic robustness evaluation of CLIP by examining multiple robustness dimensions, including visual factor variations, uncertainty calibration, out-of-distribution detection, and modality interaction.
However, their focus remains largely classification-centric.
Other analysis work has investigated limitations in basic image augmentations comprehension for CLIP and SigLIP, revealing gaps in image-level understanding under transformations~\cite{anis2025limitations}. More recently, studies have also shown brittleness in CLIP’s text encoders under lexical, syntactic, and semantic perturbations, pointing to robustness issues beyond the visual modality~\cite{tran2025brittleness}. Despite these advances, robustness under natural adversarial scenarios and evaluations beyond image classification remain underexplored. Our work addresses this gap by evaluating natural adversarial robustness across multiple settings and tasks, comparing CLIP, robustness-enhanced CLIP variants, SigLIP2, and BLIP2.

\section{Evaluation Framework}
\begin{figure}
\includegraphics[width=\textwidth]{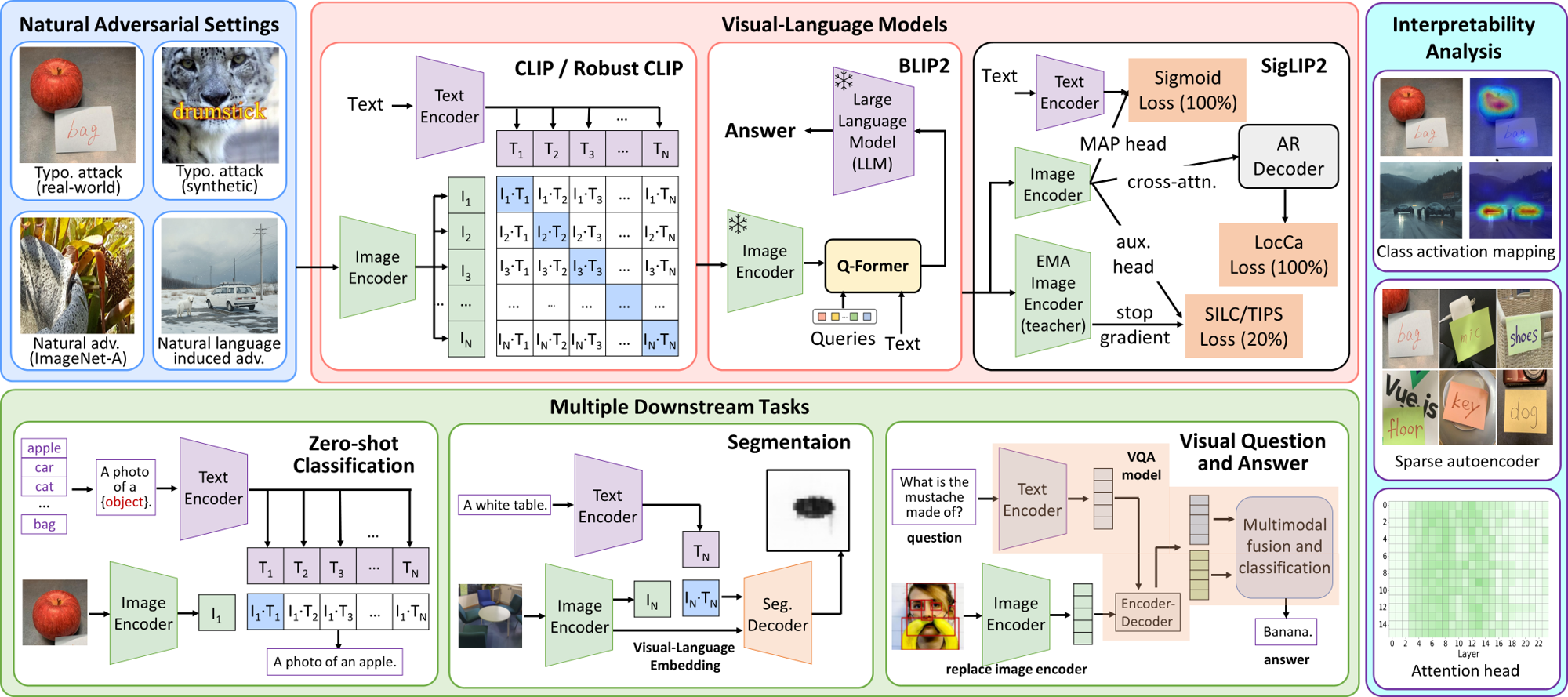}
\caption{Proposed evaluation framework. Vision-language models are evaluated against typographic attacks, ImageNet-A, and natural language-induced adversarial examples across multiple downstream tasks. Framework also supports interpretability analysis.} \label{fig:framework}
\end{figure}

We propose a comprehensive natural adversarial evaluation framework for vision-language pretrained image encoders (CLIP-style models). Figure~\ref{fig:framework} provides an overview of the proposed framework, which systematically analyzes the robustness and interpretability of CLIP-style models under diverse natural adversarial scenarios across various downstream tasks.

\vspace{-0.5cm}

\vspace{-0.5cm}\subsection{Targets of Evaluation}
Our evaluation targets a set of representative image encoders drawn from modern vision–language models, selected to reflect diverse training objectives, architectural designs, and robustness strategies. Specifically, our evaluation encompasses 22 vision–language models spanning seven families: 3 OpenAI CLIP~\cite{radford2021learning}, 1 timm ResNet CLIP~\cite{rw2019timm}, 6 Robust CLIP~\cite{schlarmann2024robust} (4 FARE \& 2 TeCoA), 4 SigLIP2~\cite{tschannen2025siglip}, 4 PaLiGemma2 ( SigLIP-So400m with Gemma2), and 4 BLIP2~\cite{li2023blip} (with T5/OPT).
CLIP and its robustness-enhanced variants serve as contrastive vision–language baselines, enabling direct comparison between standard pretraining and adversarially fine-tuned representations. SigLIP2 represents large-scale contrastive models trained with alternative loss formulations and data curation strategies, offering a strong non-CLIP contrastive baseline. BLIP2 introduces a different multimodal design that combines pretrained vision encoders with lightweight language models, allowing us to examine robustness under a hybrid discriminative–generative paradigm.
Across all model families, we focus on the image encoders as the primary evaluation target. All models are evaluated under a unified protocol across the same downstream tasks and natural adversarial settings.

\subsection{Natural Adversarial Scenarios}
We construct three types of natural adversarial scenarios to evaluate model robustness in realistic settings: (1) typographic attacks with visually confusing text, (2) natural adversarial examples that are inherently ambiguous or misclassified, and (3) natural language-induced adversarial examples, where prompts (that carry adversarial semantic information) mislead the model. These scenarios aim to reflect real-world challenges beyond synthetic perturbations.

\vspace{2mm}\noindent{\bf Typographic Attacks.}
To evaluate robustness against visually deceptive textual cues, we construct a typographic attack dataset comprising both real-world and synthetic settings. Real-world examples were collected from datasets such as the real-world typographic attack dataset (RTA-100)~\cite{azuma2023defense}, which features 1,000 images across 100 categories with physically attached misleading text (\ie, printed or handwritten labels like stickers). Synthetic examples are generated by overlaying text (\ie, incorrect class names) onto images using electronic fonts. 

\vspace{2mm}\noindent{\bf Natural Adversarial Examples.}
We use ImageNet-A~\cite{hendrycks2021natural}, a challenging benchmark of natural adversarial examples created through adversarial filtration, which are collected from various sources (iNaturalist, Flickr) related to 200 ImageNet classes, and we apply a rigorous filtering process: images are retained only if ResNet-50 models consistently misclassified them with low confidence (confidence in correct class $<15\%$). This process can be formalized as selecting images X where $\max(p_\theta(y_{\text{true}}|x)) < t$ for all model variants $\theta$ in the filtering ensemble, with threshold $t=0.15$. The resulting dataset contains 7,500 naturally occurring images that cause consistent failures across models despite being easily recognizable by humans ($\approx90\%$ human accuracy).

\vspace{2mm}\noindent{\bf Natural Language-Induced Adversarial Examples.}
A recent study showed that some high-frequency semantic information such as ``foggy'', ``humid'', ``stretching'', \etc in real images can cause classification errors (known as natural language-induced adversarial examples~\cite{zhu2024natural}).
We construct a natural language-induced adversarial dataset for 10-class animal classification by optimizing discrete text prompts $p$ (\eg, ``two green dogs wearing clothes stretching on a foggy day'') using an adaptive genetic algorithm to maximize the attack success rate as in~\cite{zhu2024natural},
\begin{equation}
\operatorname{ASR}(p)=\frac{\#\{f(G(p))\neq y\}}{\#\{G(p)\}},
\end{equation}
and semantic consistency,
\begin{equation}
\operatorname{SEM}(p)=\operatorname{CLIP}(G(p),g_t),
\end{equation}
where $G$ is Z-Image~\cite{team2025zimage} (open-source high-efficiency image generation model), $f$ is the target classifier, $y$ is a true class, and $g_t$ is true class prompt (\eg, ``a photo of a cat''). The prompts, initialized with 20 variants per animal, evolve over 8 generations via crossover/mutation ($pm=0.01$) and word-space pruning, yielding 160 adversarial images per class with high transferability $(\operatorname{ASR}(p) > 84\%)$.

\subsection{Evaluation Tasks}
\noindent{\bf Image Classification.}
We evaluate zero-shot image classification performance following the protocol established by Radford et al. \cite{radford2021learning}. For each category, class labels are embedded into textual queries using the descriptive template: ``a photo of a {class label}.''
We additionally incorporate the learned logit scale and bias parameters during the similarity calculation to align with its sigmoid loss training objective for SigLIP2 due to its architectural divergence and lack of a native prompt-based classification head.
We exclude BLIP2 from this evaluation because it is not designed for prompt-based zero-shot classification.
We use ImageNet-1K as the clean dataset. Natural adversarial variants included: (1) typographic attacks, with both RTA-100~\cite{goh2021multimodal} (real-word replacements) and ImageNet-typo~\cite{ilharco2022patching} (distracting digital text with confusing labels added to ImageNet-1K images), (2) ImageNet-A~\cite{hendrycks2021natural}, a collection of naturally adversarial examples, and (3) a natural language-included adversarial dataset~\cite{zhu2024natural}, built from Animal-10 by selecting 100 images per class (1,000 total), where misleading textual elements are embedded.

\vspace{2mm}\noindent{\bf Semantic Segmentation.}
For CLIP-based models, we attach a CLIPSeg decoder~\cite{luddecke2022image} to the CLIP vision encoder, injecting selected intermediate transformer features via skip connections during inference. For SigLIP2, we adapt its vision encoder using a lightweight segmentation head that operates on intermediate visual features, following standard encoder–decoder designs for dense prediction. For BLIP-2, we adapt its frozen visual backbone by attaching a lightweight decoder. Selected intermediate visual features are injected into the decoder via skip connections, following an effective design for dense prediction.
We use PhraseCut \cite{wu2020phrasecut} as the clean dataset. Adversarial versions included: (1) a synthetic typographic variant with added confusing text overlays and (2) segmentation annotations on ImageNet-A and the Animal-10 adversarial set using Grounded-Segment-Anything~\cite{carion2025sam3segmentconcepts}.

\vspace{2mm}\noindent{\bf Visual Question Answering (VQA).}
We use CLIP-ViL \cite{shenmuch} for all CLIP models, retaining the MCAN~\cite{yu2019deep} setup (LSTM encoder and modular co-attention over grid features extracted via reshaped ViT patches or pre-pooling ResNet layers). PaLiGemma2 with SigLIP-So400m and Gemma2 text encoders for SigLIP2 models, and BLIP-2 models were evaluated using few-shot VQA for blip2-opt-2.7b and zero-shot settings for the remaining variants. We use the VQA-v2 dataset \cite{antol2015vqa} as the clean benchmark. Adversarial sets included: (1) a synthetic version of VQA-v2 with typographic perturbations in the image and (2) question-answer pairs automatically generated over ImageNet-A and Animal-10, produced using GPT-4o. Hereinafter, (LangAdv) denotes natural language-induced adversarial examples, and (-typo.) denotes VQA-v2 with synthetic typographical attack.

\subsection{Evaluation Metrics} 
We use standard evaluation metrics for each downstream task. For classification, we report top-1 accuracy. For VQA, we evaluate using accuracy and exact match. For segmentation, we report two metrics: foreground intersection-over-union (IoU\textsubscript{FG}), which measures the overlap between predicted and ground-truth foreground regions, and pixel accuracy, which calculates the percentage of correctly predicted pixels over the entire image.

\subsection{Interpretability Analysis}
To gain insight into the vision encoder’s behavior under adversarial conditions, we employ three complementary interpretability techniques: class activation mapping (CAM)~\cite{chefer2021generic}, sparse autoencoder-based feature analysis (SAE)~\cite{limsparse}, and attention head visualization~\cite{zheng2025spot}. These methods respectively highlight salient input regions, uncover latent semantic structure, and examine attention distribution patterns, offering a multi-faceted understanding of model vulnerabilities.

\vspace{2mm}\noindent{\bf Class Activation Mapping.}
To investigate the spatial sensitivity of the vision encoder, we apply a generalized CAM method tailored for Transformer architectures. Unlike traditional CAM approaches designed for CNNs, it leverages attention-based relevance propagation to construct class-specific heatmaps that highlight regions contributing most to the model's output. Let $A\in \mathbb{R}^{h\times q\times k}$ denote the attention map from a multi-head self-attention layer, where $h$ is the number of heads, $q$ is the number of query tokens, and $k$ the number of key tokens. Following Chefer et al.~\cite{chefer2021generic}, we compute a head-aggregated attention relevance map $\overline{A}\in \mathbb{R}^{q\times k}$ using gradient-based weighting:
\begin{equation}
    \overline{A}=\mathbb{E}_h((\nabla A\circ A)^+),
\end{equation}
where $\nabla A=\frac{\partial{y^t}}{\partial{x}}$ is the gradient of the model output for class $t$, the operator, ($\circ$) denotes the Hadamard product, and $(\cdot)^+$ clips negative contributions. Relevance is propagated across layers using update rules:
\begin{equation}
    R_{qq} \leftarrow R_{qq}+\overline{A}\cdot R_{qq},\enspace R_{qk} \leftarrow R_{qk}+\overline{A}\cdot R_{qk}.
\end{equation}

\vspace{2mm}\noindent{\bf Sparse Autoencoder-based Feature Encoding.} To uncover the latent structure of the representations learned by vision encoder and to analyze its sensitivity to natural adversarial examples, we use sparse autoencoders (SAE) as a tool for post hoc interpretability. SAEs aim to disentangle polysemantic neuron activations into monosemantic units by learning a sparse dictionary over the embedding space of the model. Given a pretrained model $f:X\rightarrow \mathbb{R}^d$, we extract activation vectors $v=f_l(x)$ from layer $l$, where $x\in X$ is the input image. An SAE consists of an encoder $\phi(v)=\sigma(W^\tau_{\text{enc}}(v-b))\in \mathbb{R}^\omega$ and a decoder $\psi(\phi(v))=W^\tau_{\text{dec}}\phi(v)+b$, where $\omega$ is the latent dimension, and $\sigma$ is typically ReLU or Top-K activation enforcing sparsity. The SAE is optimized with the loss:
\begin{equation}
    L(v)=||v-\hat v||^2+\lambda||(\phi(v))||_1,
\end{equation}
where $\hat v=\psi(\phi(v))$, and $\lambda$ controls the sparsity level. To quantify the interpretability of each latent neuron, we use the Monosemanticity Score ($\operatorname{MS}$), defined for neuron $k$ over a dataset of $N$ images as:
\begin{equation}
    \operatorname{MS}_k=\frac{1}{N(N-1)}\sum_{n\neq m}r^k_{nm}s_{nm},
\end{equation}
where $s_{nm}$ is the cosine similarity between CLIP embeddings of images $x_n$ and $x_m$, and $r^k_{nm} = \overline{a}_{kn}\overline{a}_{km}$ weights the similarity by using the normalized activation values of neuron $k$. High $\operatorname{MS}$ values indicate neurons that consistently respond to semantically coherent image concepts, making them easier to interpret and localize. In the context of natural adversarial perturbations, we analyze how monosemantic neurons change their activation patterns.

\vspace{2mm}\noindent{\bf Attention Head.} To investigate model's vulnerability to natural adversarial images, we analyze attention heads in vision encoder, hypothesizing that specific heads encode safety-related features based on recent LVLM findings~\cite{zheng2025spot}. Formally, for an input image $x$, the vision encoder processes a tokenized input sequence through multiple Transformer layers, each with $H$ attention heads. In the l-th layer, the attention output from head $h$ is denoted as:
\begin{equation}
 a^h_l=\operatorname{Attention}^h_l(Q^h_l,K^h_l,V^h_l),
\end{equation}
where $Q$, $K$, and $V$ are query, key, and value matrices derived from input token embeddings. The final output of the layer is:
\begin{equation}
 x_{l+1}=x_l+\sum^H_{h=1}O^h_la^h_l,
\end{equation}
with $O^h_l$ as the output projection for head $h$.

\section{Evaluation Results \& Analysis}

\subsection{Zero-Shot Image Classification}
\begin{figure}
\includegraphics[width=\textwidth]{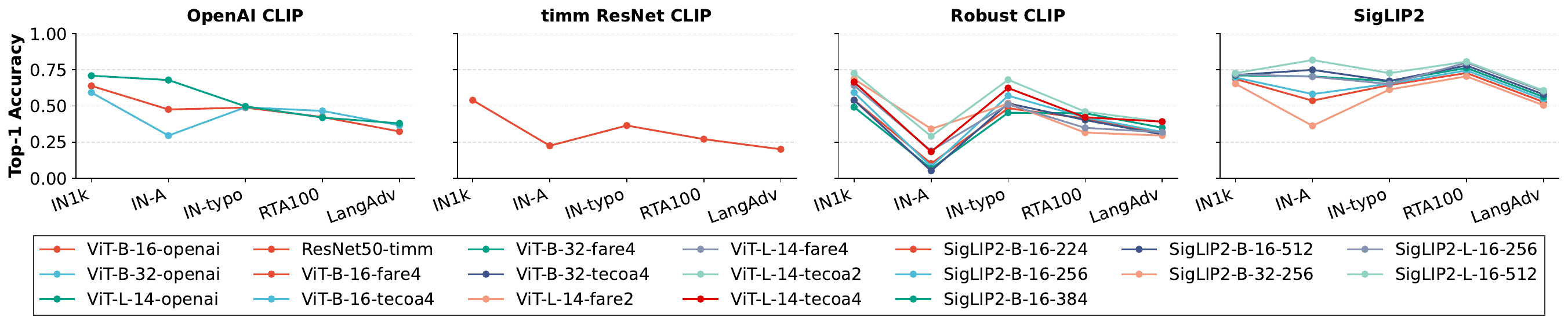}
\caption{The classification performance across clean and natural adversarial datasets. IN: ImageNet. IN-A: ImageNet-A. IN-typo: ImageNet-typographic. LangAdv: language induced adversarial images.} \label{fig:classification}
\end{figure}

As shown in Fig.~\ref{fig:classification}, CLIP’s performance drops significantly under natural adversarial conditions, particularly on LangAdv (38.25\%). Interestingly, despite being designed for robustness, robust CLIP suffers from even lower accuracy on ImageNet-A (34.17\%) and RTA-100 (31.6\%). In contrast, SigLIP2 achieves superior robustness across nearly all scenarios, suggesting that its training regimen-potentially involving more diverse or targeted data mixtures-effectively addresses challenging out-of-distribution shifts and semantic perturbations.

\vspace{0.1cm}

\subsection{Semantic Segmentation}
\begin{figure}
\includegraphics[width=\textwidth]{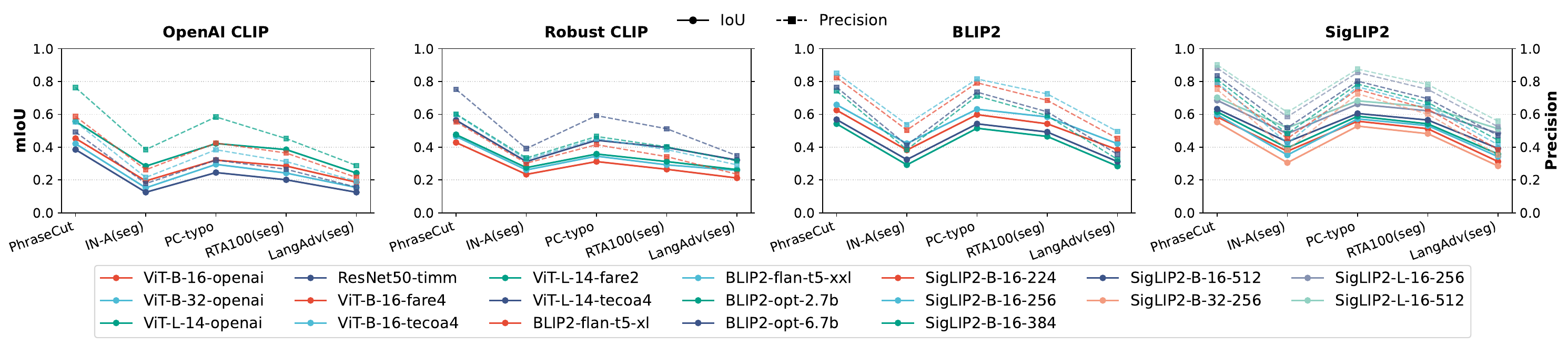}
\caption{The segmentation performance across clean and natural adversarial datasets. PC-typo: PhraseCut-typographic. IN-A: ImageNet-A. LangAdv: language induced adversarial images.} \label{fig:segment}
\end{figure}

Figure~\ref{fig:segment} reports the average foreground IoU and precision for segmentation under clean and natural adversarial conditions. While all models achieve reasonable performance on PhraseCut, segmentation quality degrades substantially under natural language adversarial prompts and synthetic typographic perturbations, highlighting the vulnerability of segmentation to both linguistic and low-level visual disturbances. 
SigLIP2 consistently outperforms other models in terms of robustness, retaining relatively higher IoU and precision across perturbations. BLIP2 is comparatively insensitive to textual interference; however, its overall segmentation performance remains poor, particularly on ImageNet-A and LangAdv, indicating limited robustness to distribution shifts.

\vspace{-0.5cm}

\vspace{-0.5cm}\subsection{Visual Question Answering}
\begin{figure}
\vspace{-0.5cm}
\includegraphics[width=\textwidth]{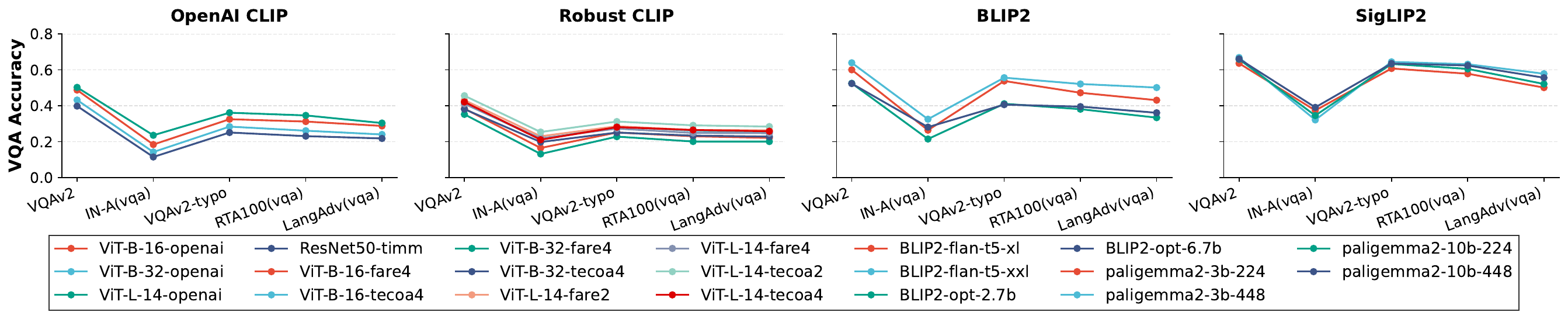}
\caption{The visual question answer performance across clean and natural adversarial datasets. IN-A: ImageNet-A. LangAdv: language induced adversarial images.} \label{fig:vqa}
\vspace{-0.5cm}
\end{figure}

Figure~\ref{fig:vqa} reports VQA performance under clean and adversarial settings. While accuracy consistently drops under adversarial perturbations, SigLIP2 remains comparatively stable across conditions. Among BLIP2 variants, FLAN-T5–based models show stronger robustness to typographic and language-based adversaries, whereas BLIP2-OPT degrades markedly under natural language attacks. Overall, these results suggest that recent models improve robustness in VQA, but adversarial sensitivity remains significant.

\subsection{Key Findings}
Across all tasks, natural language-induced adversarial prompts consistently result in the most substantial performance drop, underscoring the inherent vulnerability of CLIP's cross-modal alignment to subtle linguistic variation. This suggests that robustness in multimodal models cannot be solely achieved through visual-domain regularization or augmentation. Furthermore, we observe that robustness-enhanced variants, while partially mitigating specific adversarial effects, often do so at the cost of performance on standard benchmarks and fail to generalize across adversarial types and tasks.

Our key findings are as follows: (1) Robust CLIP was less robust than standard CLIP under natural adversarial scenarios, particularly with challenging or linguistically perturbed inputs. (2) Under natural adversarial conditions, CLIP's performance fell sharply in classification, segmentation, and VQA, especially with language-induced attacks. (3) SigLIP2 consistently demonstrates superior overall robustness across all evaluated tasks, suggesting more effective handling of semantic perturbations. (4) While BLIP2 partially mitigates typographic attacks, it still suffers significant performance degradation under complex perturbations such as ImageNet-A and LangAdv. (5) In segmentation and VQA tasks, variants utilizing the FLAN-T5 backbone exhibit slightly better resilience than OPT-based counterparts.

\begin{figure}
\vspace{-0.5cm}
    \includegraphics[width=\textwidth]{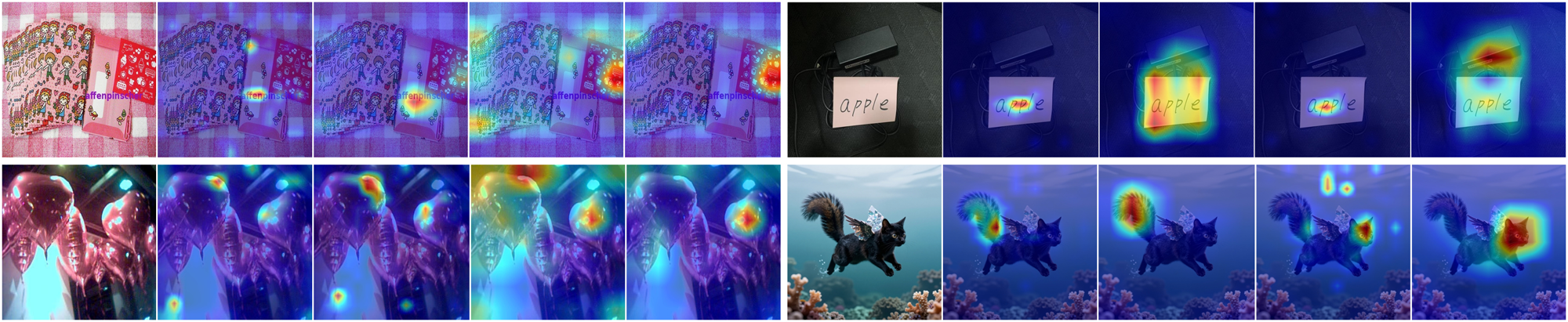}
    \caption{GradCAM of vision-language models in different natural adversarial images. Quadrants: ImageNet-Typo, RTA100 (top); ImageNet-A, LangAdv (bottom). Columns: Original, CLIP, robust CLIP, BLIP2, SigLIP2. Heatmaps denote high (red) to low (blue) attention intensity.}
    \label{fig:CAM}
\vspace{-0.5cm}
\end{figure}

\subsection{Interpretable Analysis}

\vspace{2mm}\noindent{\bf CAM.} We analyze the spatial attention behaviors using CAM visualizations (Fig.~\ref{fig:CAM}).
Under typographic perturbations, CAM results show a consistent shift of attention from the true semantic object to overlaid or embedded text regions. This effect is most pronounced in CLIP, where text dominates the attention map regardless of semantic relevance, reflecting a strong text-centric inductive bias induced by large-scale web pretraining.
Robust CLIP exhibits smoother and less concentrated attention patterns due to adversarial training; however, it remains sensitive to textual interference, indicating that improved robustness stabilizes attention but does not fundamentally resolve text over-reliance.
BLIP2 and SigLIP2 display weaker attention bias toward text when object semantics are clear, but still exploit textual cues when visual features are ambiguous, suggesting that text acts as a secondary signal under visual uncertainty. 

On ImageNet-A, CLIP, Robust CLIP, and BLIP2 produce diffuse and poorly localized activations in cluttered scenes, indicating limited object localization under high visual complexity. In contrast, SigLIP2 better preserves focus on the salient object, likely benefiting from targeted training on more challenging samples.

On the LangAdv dataset, CLIP-based models and BLIP2 exhibit noticeable attention shifts toward non-essential regions, leading to inconsistent visual grounding across samples. SigLIP2, however, shows no clear feature displacement in the examined cases, suggesting stronger alignment between visual attention and semantic content.

\begin{figure}
\includegraphics[width=\textwidth]{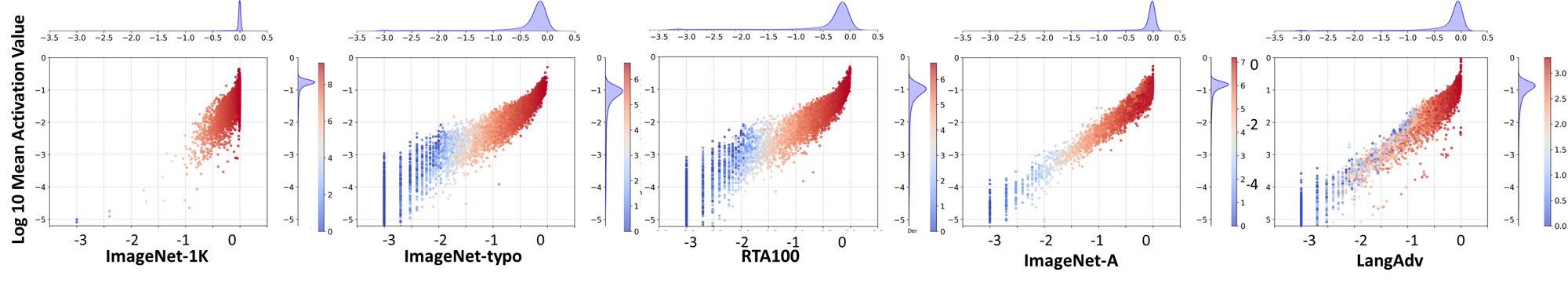}
\caption{The statistical distribution of SAE latent features across standard (ImageNet-1K) and various natural adversarial datasets. Each point represents a latent feature, positioned by its activation frequency (x-axis) and mean activation magnitude (y-axis), with color-coding indicating label entropy.} \label{fig:SAE}
\end{figure}

\vspace{-0.5cm}\noindent{\bf SAE.} The Sparse Autoencoder (SAE) latent distributions reveal a fundamental shift in vision-language model’s feature activation patterns when transitioning from standard to adversarial datasets. In ImageNet-1K, the latent space is dominated by high-magnitude, high-entropy features (red points in Fig.~\ref{fig:SAE}), indicating that the model relies on a dense set of broad, overlapping features for standard classification. In contrast, natural adversarial datasets such as RTA100, ImageNet-A, and LangAdv exhibit a significant emergence of low-entropy, sparsely activated features (blue points in Fig.~\ref{fig:SAE}) in the lower-frequency regions.

This suggests that adversarial perturbations (such as typographic noise or complex backgrounds) force the model to activate ``atypical'' or highly specific latents that are normally dormant or suppressed. The performance degradation on these datasets can thus be explained as a feature-level fragmentation: the adversarial inputs distract the model from its primary high-confidence features, triggering a long tail of specialized but less robust latents that lead to inconsistent and incorrect semantic mappings.

\begin{figure}
    \vspace{-0.5cm}
    \includegraphics[width=1.0\textwidth]{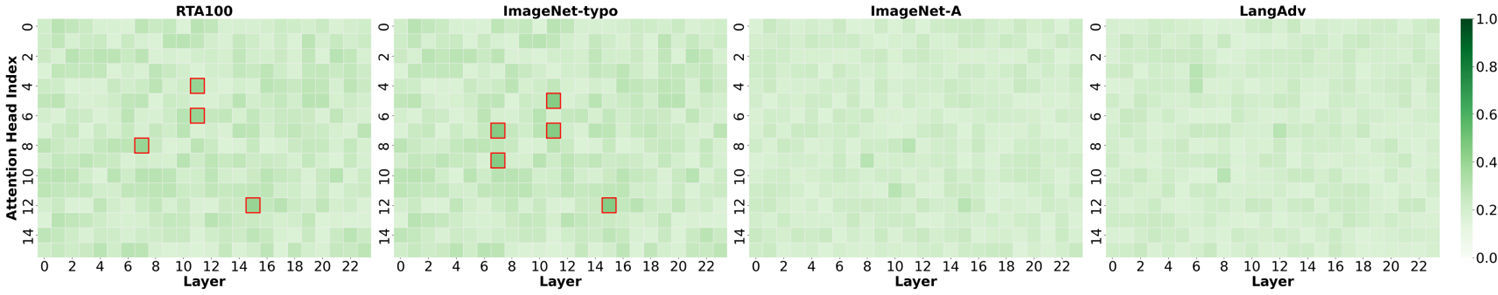}
    \caption{Accuracy variation of CLIP via single-head masking on natural adversarial examples. Highlighted areas denote critical heads where masking induces a substantial accuracy (>15\%). \label{fig:attention}}
    \vspace{-0.5cm}  
\end{figure}

\vspace{2mm}\noindent{\bf Attention Head.} Visualization of attention heads further reveal modality-specific vulnerabilities, as shown in Fig.~\ref{fig:attention}. In typographic attack images, certain attention heads within the visual transformer layers are disproportionately activated by textual patterns, often leading to incorrect or overly confident predictions. These heads display behavior analogous to adversarially triggered neurons in standard deep networks, highlighting potential structural fragility. For ImageNet-A and natural language-induced adversarial examples, we do not observe consistent or discriminative head activation patterns, suggesting that the failure in these scenarios is more diffuse and stems from high-level misalignment rather than localized structural bias. This limits the efficacy of targeted pruning or reweighting strategies in such cases. In contrast, BLIP-2 and SigLIP-2 do not exhibit pronounced head-wise biases across the four adversarial settings, indicating a more resilient and distributed representation mechanism.

\section{Discussion}
Our evaluation demonstrates that SigLIP2 exhibits consistently stronger robustness under natural adversarial settings, preserving semantic alignment and object-centric attention in cluttered or perturbed scenes. In contrast, Robust CLIP frequently underperforms standard CLIP, suggesting that adversarial fine-tuning tailored to synthetic perturbations does not generalize well to naturally occurring adversarial variations. Interpretable analyses indicate that CLIP’s failures largely stem from a persistent over-reliance on local textual artifacts, and that adversarial fine-tuning can degrade cross-task generalization and multimodal reasoning. To address these issues, we recommend incorporating more diverse and semantically challenging training data, as well as targeted interventions such as contrastive adversarial training and attention head pruning, while carefully balancing robustness gains against performance on clean data.

\vspace{2mm}\noindent{\bf Limitations.} We primarily assess robustness through frozen image encoders with task-specific adapters, which may not fully capture the behavior of end-to-end vision-language systems. In addition, while our natural adversarial benchmarks span multiple settings, they do not exhaust the full space of real-world adversarial conditions. Finally, in challenging datasets such as ImageNet-A, failure modes are often diffuse and non-discriminative, making it difficult for the evaluation to attribute errors to specific architectural components or localized mechanisms.

\vspace{2mm}\noindent{\bf Future Research Directions.} (1) Future research should rethink the efficacy of adversarial fine-tuning, as strategies tailored for synthetic perturbations may inadvertently amplify vulnerabilities to natural adversarial variations. (2) Beyond visual-domain regularization, future designs must address the fragility of cross-modal alignment under subtle linguistic variations to achieve holistic robustness. (3) Targeted structural interventions, such as pruning safety-aware attention heads, should be explored to mitigate the model's persistent over-reliance on local textual artifacts. (4) Developing more resilient models requires moving beyond standard benchmarks to incorporate diverse and semantically challenging data mixtures that better represent real-world distribution shifts.

\section{Conclusion}
This paper provides the first holistic evaluation of vision-language models across classification, segmentation, and visual question answering under natural adversarial scenarios. Our results show that robust CLIP underperforms CLIP in most cases, revealing its limited generalization. We further identified key failure patterns via interpretable analysis and demonstrated that targeted strategies-such as contrastive adversarial training and attention head pruning-can improve robustness, though often with trade-offs. Our findings highlight the need for more adaptive and generalizable robustness solutions for image encoders.


%
%
%
\bibliographystyle{splncs04}
\bibliography{refer}
%





\end{document}